\documentclass{llncs}
\usepackage{url}
\makeatletter
\def\url@nitinstyle{%
  \@ifundefined{selectfont}{\def\UrlFont{\sf}}{\def\UrlFont{\footnotesize\ttfamily}}}
\makeatother
\usepackage[T1]{fontenc}
\usepackage{subfigure}
\usepackage[pdftex]{color,graphicx}

\title{Active Learning for Mention Detection: A Comparison of Sentence Selection Strategies}
\author{Nitin Madnani$^{a}$, Hongyan Jing$^{b}$, Nanda Kambhatla$^{b}$ \& Salim Roukos$^{b}$\\ \texttt{nmadnani@umiacs.umd.edu} \quad \texttt{\{hjing,nanda,roukos\}@us.ibm.com}}
\institute{$^a$Department of Computer Science, University of Maryland College Park\\$^b$IBM T.J. Watson Research Center, Yorktown Heights, NY}
\date{}

\begin{document}
\maketitle
\begin{abstract}

We propose and compare various sentence
selection strategies for active learning for the task of detecting mentions
of entities. The best strategy employs the sum of confidences of two statistical classifiers trained on different views of the data. Our experimental results show that, compared to the random
selection strategy, this strategy reduces the
amount of required labeled training data by over 50\% while achieving
the same performance.  The effect is even more significant when only
named mentions are considered: the system achieves the same
performance by using only 42\% of the training data required by the
random selection strategy.

\end{abstract}

\section{Introduction}

Human annotation is expensive, yet it is needed in many tasks
in order to create training data. Given the high cost, it is critical to improve the efficiency of such annotation.  Active learning
involves selecting samples ``intelligently'' rather than randomly, for human annotation. By doing so, it is possible for systems to attain
better performance with the same amount of annotation or achieve the same level of performance with a lot less annotated data. 

In this paper, we present several new active learning strategies for the task of Mention Detection (MD). Here, we employ the terminology used in the
Automatic Content Extraction Conferences~\cite{ACE}. Mentions are
references to real-world entities that can be \emph{named}
(e.g. ``John''), \emph{nominal} (e.g. ``survivor'') or
\emph{pronominal} (e.g. ``he''). 

We propose and investigate a variety of sentence selection criteria for
active learning, including various sentence scoring metrics that
combine uncertainty-based and query-by-committee like
measurements.  Experimental results show that these sentence selection
strategies are quite effective for mention detection: compared to the random
selection strategy, the best strategy reduces the amount
of required annotated training data by over 50\% while achieving the
same performance. The effect is even more significant when only named mentions are considered: the system achieves the same performance by using only 42\% of the training data required by the random strategy.

In the next section, we discuss related work on active learning.
Section~\ref{setup} describes our framework in detail and presents our
experimental setup. Section~\ref{results} presents the results of
the different experiments. Finally, we discuss observations from our
experiments and present some ideas for future work.

\section{Related Work}\label{relatedwork}

Active learning has been utilized for many NLP applications, most
noticeably for text
classification~\cite{Lewis&Catett1994,McCallum&Nigam1998,Schohn&Cohn2000,Tong&Koller2001}. It
has also been applied for part-of-speech
tagging~\cite{Engelson&Dagan1995}, statistical
parsing~\cite{Thompson&al1999,Tang&al2002,Steedman&al2003,Osborne&Baldridge},
noun phrase chunking~\cite{Ngai&Yarowsky2000}, Japanese word
segmentation~\cite{Sassano2002}, and confusion set
disambiguation~\cite{Banko&Brill}.

There are few reported instances of applying active learning
techniques to mention detection.  \cite{Finn&Kushmerick} investigated
several document, rather than sentence, selection strategies for Information Extraction.
They observed that some strategies lead to improvement
in recall while others improve precision, but it is difficult to get
significant improvement in both recall and precision for an active
learner to perform better than random selection.

\cite{Shen&al} proposed a multi-criteria-based active learning
approach for named entity recognition. The multiple criteria include informativeness,
representativeness, and diversity. They proposed measures to quantify
these properties and investigated different selection strategies. They
showed that the labeling cost can be reduced by at least 80\% without
degrading the performance for their data sets.

\cite{Hachey&al2005} investigated a query-by-committee-based active
learner for information extraction in the astronomy domain. It studied
the effects of selective sampling on human annotators. Although active
learning improved annotation efficiency overall, they observed lower
inter-annotator agreement and higher per-token annotation times for
the data selected by active learning.

\section{Active Learning}\label{setup}

Active learning techniques are usually divided into two types:
uncertainty sampling for a single learner~\cite{Cohn&al1995}, or disagreement measurement between
a committee of learners~\cite{Seung&al1992}. In each case, seed data needs to be provided
to build an initial model or models. In the uncertainly-based approach,
a single learner labels unlabeled examples and provides a confidence score
for each predicted label. Samples that have the lowest confidence scores are
chosen for manual labeling. In the query-by-committee approach, a committee
of learners is built and each learner labels the unlabeled
samples. Samples that have the highest disagreement among committee
members are chosen for manual labeling.

In our work, we experiment with query-by-committee-based approaches
as well as hybrid approaches in which we employ a weighted combination of multiple learners. We also explore the effect of using
different sets of committee learners. We carry out a number of
experiments to compare the selection strategies that we propose. In this
section, we first describe the corpus used in the experiments and then
present the statistical classifiers used in the committees, along with the scoring metrics that we use.

\subsection{The MALACH Information Extraction Corpus}

The MALACH
collection~\cite{malach_jcdl_2002,oard04:_build_infor_retriev_test_collec}
contains 116,000 hours of digitized interviews and testimonies in 32 languages from
52,000 survivors, liberators, rescuers and witnesses of the Holocaust. We are interested in automatic information extraction from this corpus of spontaneous conversational speech. For a small number of English testimonies, we manually transcribed them and manually annotated the transcripts with three types of information:

\begin{itemize}

\item \textbf{Mentions}. Named, nominal and pronominal mentions of 20 categories of entities.

\item \textbf{Co-reference}. Sets of mentions that refer to the same real-world entity.

\item \textbf{Relations}. Sets of relationships between pairs of mentions.  For instance, given the sentence ``I was in Auschwitz
for a year'', the following relation exists: \texttt{LocatedAt(I, Auschwitz)}.

\end{itemize}

For the experiments described in this paper, we chose to focus on mention
annotation only. We plan to explore co-reference and relation annotation in the future. For mention
annotation, we excluded pronouns; their inclusion leads to inflated numbers since detection of pronominal mentions is easy and they occur with high frequency.

We split the annotated data into two parts: the pool from which the
learners select the next batch of data for annotation, and the development
set. The first pool consists of 99 documents, including 4772
sentences and 198K words. A total of 43K mentions have been annotated
for the documents in the pool (16K mentions if pronouns are
excluded). The development evaluation set consists of 5 complete
testimonies, which cover over 10 hours of speech. It consists of
1700 sentences, 73K words and 16K total mentions (6K mentions if pronouns are excluded). 


\subsection {Granularity of Selection}

When using active learning techniques to select samples for human
annotation, we need to first decide on the granularity of sample
selection. The samples can be documents, sentences, or
tokens. A possible problem with selecting samples at the document
level \cite{Finn&Kushmerick} is that a document may only be partially useful from a learning point of view and and it
is impossible to add only the interesting examples so as to maximize the effect of active learning. This problem is particularly acute in our
corpus since the documents, which are testimonies of survivors, are
very long and contain a lot of redundant information. Selecting tokens as samples, as in \cite{Shen&al}, has the advantage that all samples are of equal length but suffers from the problem that a human annotator has to annotate mentions by looking at only the tokens. The selected tokens may also contain partial mentions when mention boundaries are incorrectly identified. Sentences, on the other hand, contain enough non-redundant contextual information for effective annotation. Therefore, we use sentence-level blocks for active learning, as in \cite{Hachey&al2005}.

\subsection{Framework Description}\label{framework}
\begin{figure}
\begin{center}
\includegraphics[scale=0.75]{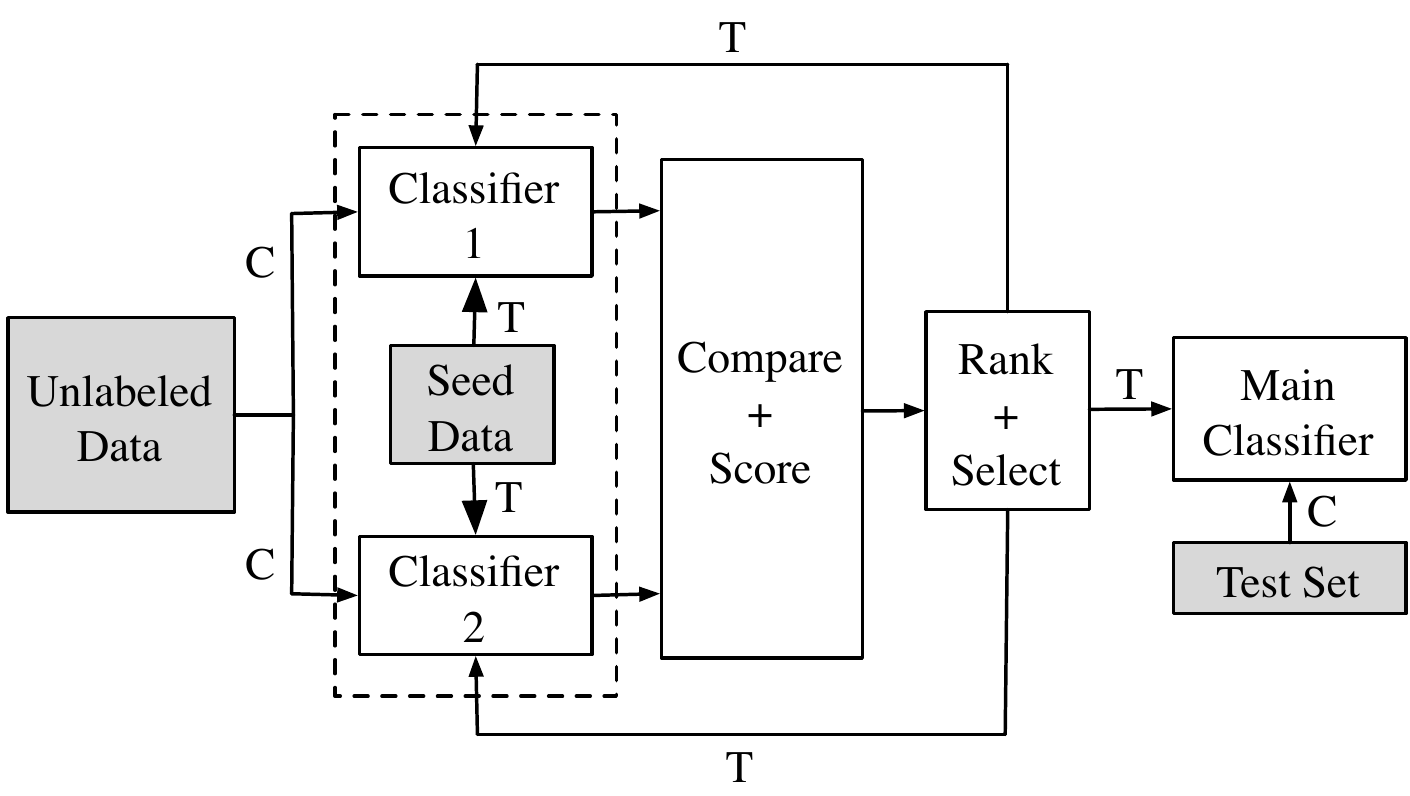}
\caption{The architecture of the active learning framework. The edges marked with a \textbf{T} represent training steps and the ones marked with a \textbf{C} represent classification.}
\label{fig:framework}
\end{center}
\end{figure}

We propose a new active learning based framework for sample selection which provides considerable savings in the annotation task and can provide  better performance for the same amount of annotation. Figure~\ref{fig:framework} shows the architecture of the framework. As mentioned before, the central idea is to use an ensemble of classifiers -each one differing from the others in some respect. Once trained, the classifiers can then be used to detect mentions in the unlabeled sentences. These detected mentions can then be compared, for each sentence, and a score assigned to this sentence indicating the agreement, or lack thereof, among the ensemble classifiers. The sentences on which the classifiers disagree the most are, intuitively, the ones that can contribute the most to the learning. The sentences are then ranked according to this metric and the required number sampled from this ranked list. \\

In the case of true active learning, these samples would be provided to the annotator to label and then added to the training data for the main classifier used for the actual task of mention detection. However, for this paper, we consider controlled experiments where the entire dataset has been annotated in advance. Therefore, the path between the selection of the samples to the annotator has effectively been short-circuited and the samples are added directly to the training data for the main classifier. We structured our setup in such a way since we wanted to experiment with adding successively increasing amounts of training data without incurring the overhead of annotation at each step. The same exact advantages and savings will apply to a real setting where the selected samples need to be annotated at each step of learning. At each step, the set of sampled sentences is split into equal parts and added to the seed data for the ensemble classifiers as well. \\

Our approach differs from previous approaches to the same problem in two significant ways. First, we measure the disagreement at the granularity of sentences rather than documents, which allows us to be much more discerning when selecting samples. The second point is that our framework allows targeting the sample selection towards specific types of mention that can be specified by the user. This is important because for some tasks, like ACE, some types of mentions carry more weight than the others. 

\subsection {Statistical classifiers}

At each step in the active learning process, we build \textbf{two} maximum-entropy based statistical classifiers~\cite{maxent-paper} using the labeled data available at that step. There are two dimensions of classifier training - the feature set of the classifier and the data used for training. For our experiments, we use two different combinations of these dimensions: 
\begin{itemize}
  \item Each classifier is trained with the \textit{same} features but on a \textit{different} half of the available labeled data. We refer to this as the \emph{data-different} (DD) setting. 
  \item Each classifier is trained on the entire available labeled data but using a \textit{different} feature set : the ``inside'' classifier uses lexical features derived solely from the current token, and the ``outside'' classifier uses features involving surrounding tokens only. We refer to this experimental setting as \emph{feature-different} (FD). 
\end{itemize}

\subsection {Sentence scoring metrics}

We employ a variety of metrics for measuring the degree of
disagreement between two classifiers over the output labels for a given
sentence. The simplest metrics look at only the output labels predicted by the classifiers. However, we propose another set of metrics that utilizes the confidence values associated with those labels as well.

\begin{itemize}

  \item \textbf{F-measure} As the harmonic mean of precision and
  recall, the F-measure is often used to assess the agreement between two
  classifiers for the named entity recognition task. The value of F-measure is between 0
  and 1, with higher values indicating greater agreement. During sentence
  selection, we compute the F-measure of the two classifiers on each
  unannotated sentence, and select sentences with the lowest
  F-measure values for annotation.

  \item \textbf{Macro-averaged F-measure} Instead of computing the
  F-measure over the entire set of mentions, as in the previous
  metric, another option is to compute the F-measure for each mention
  category and then take the average over all categories. This metric allows
  categories with a small number of mentions to be weighted
  equally. 


  \item \textbf{Confidence Sum}. Our statistical classifiers can
   provide a probability value for each output label, indicating its
   confidence in choosing the label. To leverage this information, we use the normalized sum of the
   confidence values of the two classifiers as another metric. The
   higher this value, the more confident both classifiers are about a
   particular sentence. Therefore, that sentence will provide little
   or no information if added to the training set and should be ranked
   lower for selection.  

  \item \textbf{Confidence Difference}. We also use the absolute
  difference of the two confidence values. A sentence with a higher
  confidence difference value indicates an explicit disagreement
  between the two classifiers when detecting mentions and, therefore,
  would prove to be more useful for annotation. 

\end{itemize}

In addition to the above selection metrics, there are three additional
parameters that were used in our experiments:

\begin{itemize}

\item \textbf{Minimum number of mentions per sentence}. This parameter
is used to deal with the sparse mention problem in sentence
selection. For instance, for a given sentence $S1$, if the first
classifier finds only one mention in it, and the second classifier finds
zero mentions, then the F-measure for the sentence is 0, which puts the
given sentence at the top of the selection list. In contrast, for
another sentence $S2$, if the first classifier finds 5 mentions, and
the second classifier finds 3 mentions, two of which overlap with the
output by the first classifier, then the F-measure is 50\%. Therefore,
$S2$ is erroneously ranked lower than $S1$ on the selection list. To eliminate sentences with very few
mentions, a user can set the minimal number of mentions per sentence
parameter to $N$, where $N$ is a non-negative number. Any sentence with
less than $N$ mentions is not considered as a candidate for active learning.

\item \textbf{Mention category weights}. Each mention category is
associated with a weight in the above metrics. This gives us
the flexibility to focus on certain categories during active learning. For instance, if a system performs weakly in the
ORGANIZATION category, we may want to customize the scoring metric
so that the active learner can pick up samples that are particularly
useful in improving the performance in this category. The categories with
higher weights are considered more important; the categories with zero
weights are not considered for active learning. 

\item \textbf{Mention level weights}. The mention level indicates
whether a mention is named, nominal, or pronominal. We can customize our learner to focus on a particular mention level by adjusting the weight
associated with it. 

\end{itemize}


\section{Results}\label{results}

We carried out a number of experiments to evaluate the different sentence selection strategies presented in the previous section. All the strategies perform better than random selection, but the confidence sum metric with the feature-different classifier training setting gives the greatest improvement.

Ideally, the active learners should be retrained each time a new
sample has been selected by active learning and annotated by a
human. In reality, this is rarely done due to time and computational
cost. Instead, active learners usually operate in ``batch'' mode --- a
set of samples, instead of a single sample, is selected at each step and
the active learners are retrained by adding all the samples in the
set. Given our granularity of selection, one way to define the size of this batch can be in term of number of sentences to be added at each step. However, this approach does not provide a way to differentiate between a sentence that is $100$ words long and another that is only $10$ words long; both are equally important. Therefore, we define the batch in terms of the number of words contained in the sentences. At each step,  we add just enough complete sentences that, among themselves, contain the given number of words or as close to that number as possible. In the experiments described in this section, this size is
$20000$ words.  First, a set of $20000$ words is randomly selected to
build seed models. Then at each step of active learning, an additional
$20000$ words are selected. We also describe the effect of this \emph{step size} at the end of this section. All the results presented in this section are on our
development evaluation set --- after each step of active learning, we train a
model using all the data that have been selected as of that step
(including the seed data), and the model's performance on the
development set is reported.

\subsection{Effect of different scoring metrics}

In the experiments described here, we consider only named and nominal
mentions. The weights for all mention categories are set to 1. 

\begin{figure}
\mbox{}\hspace{-0.65in}
\begin{minipage}[b]{0.6\linewidth} 
\mbox{}\hspace{-0.2in}
\includegraphics[scale=0.3]{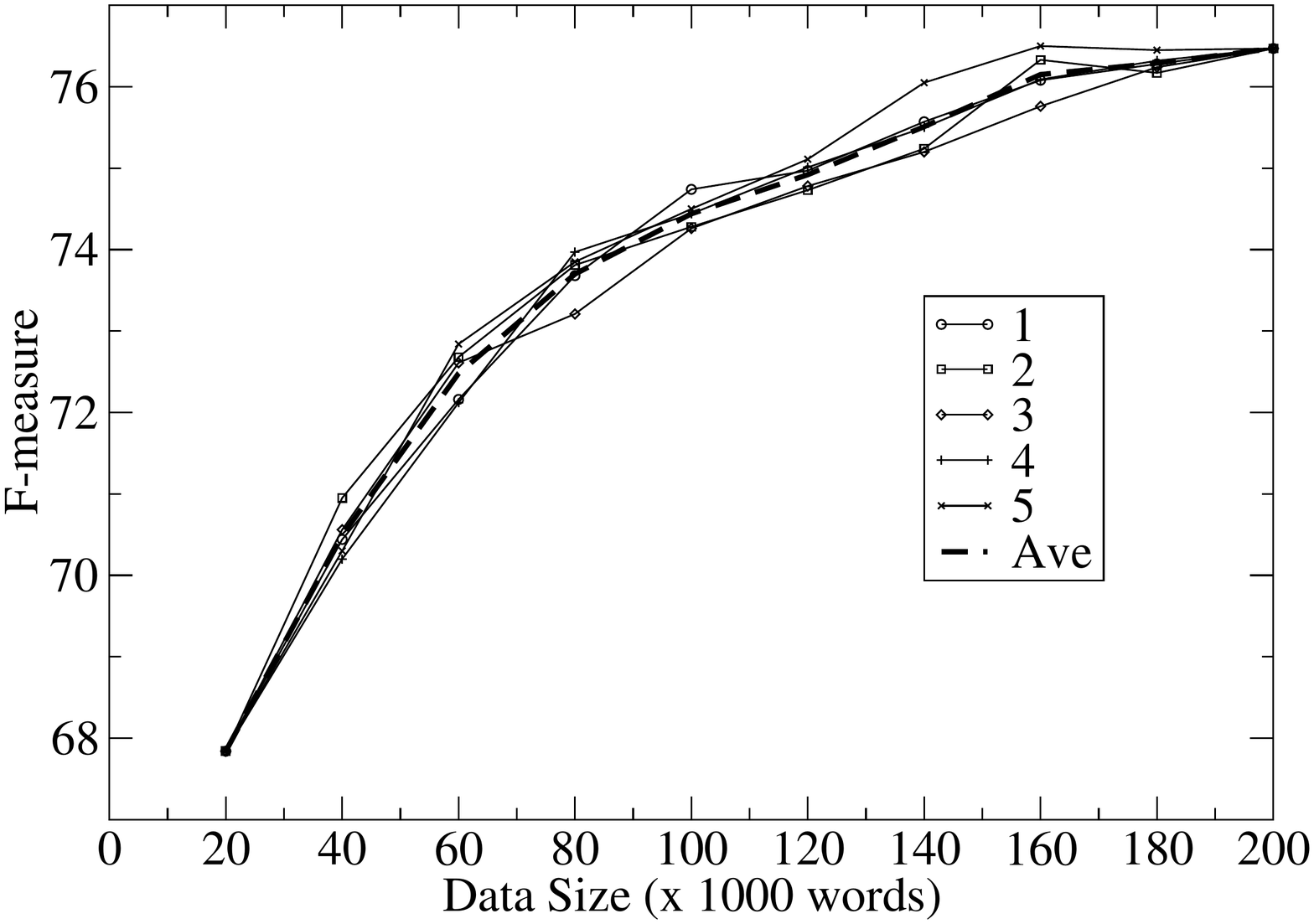}
\caption{Performance for the baseline strategy of selecting sentences at random.}
\label{baseline}
\end{minipage}
\hspace{0.1cm} 
\begin{minipage}[b]{0.6\linewidth}
\mbox{}\hspace{-0.2in}
\includegraphics[scale=0.3]{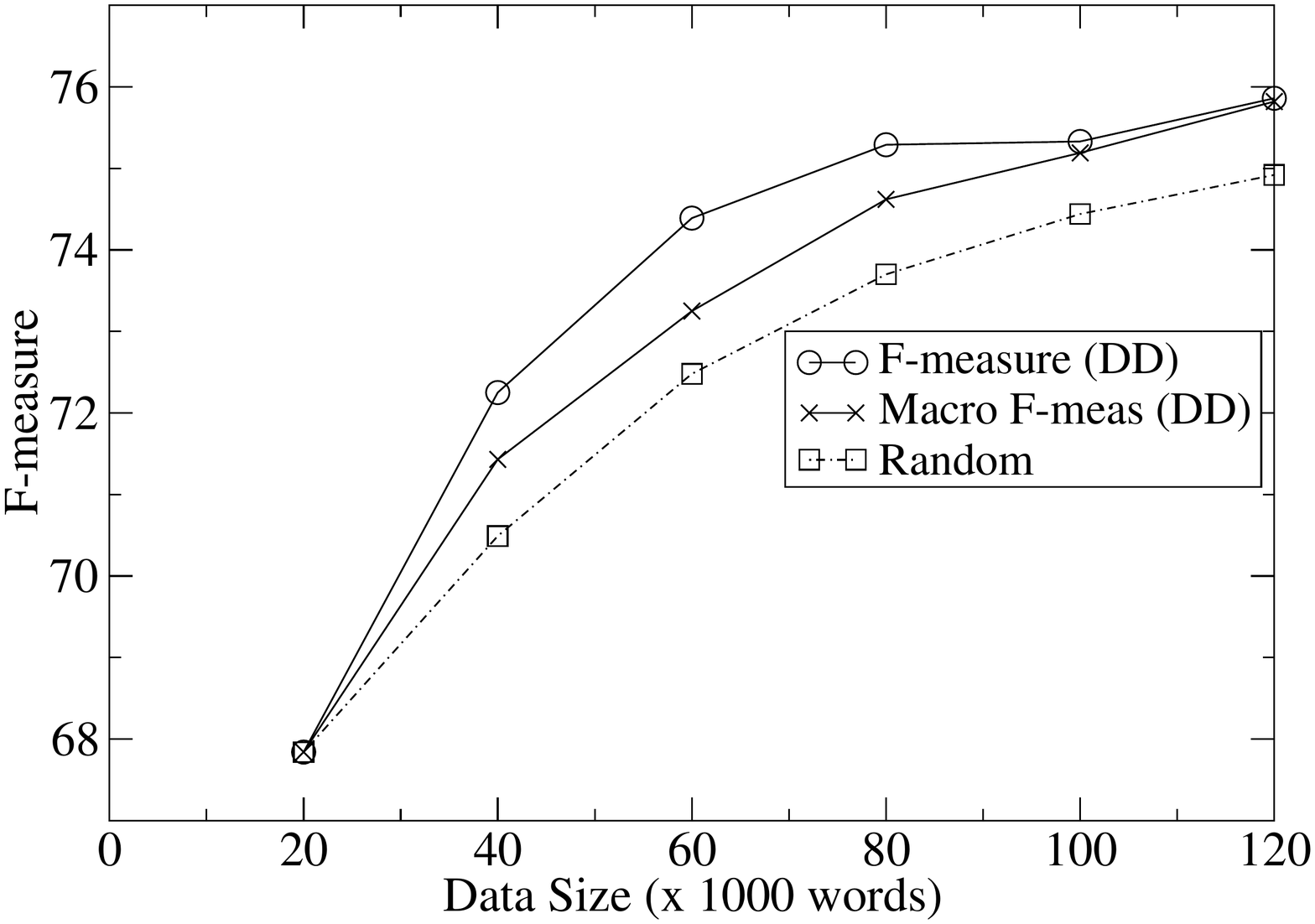}
\caption{Performance for the F-measure and Macro-Averaged F-measure metrics.}
\label{fmeas}
\end{minipage}
\end{figure}

\begin{figure}
\mbox{}\hspace{-0.65in}
\begin{minipage}[b]{0.6\linewidth} 
\mbox{}\hspace{-0.2in}
\includegraphics[scale=0.3]{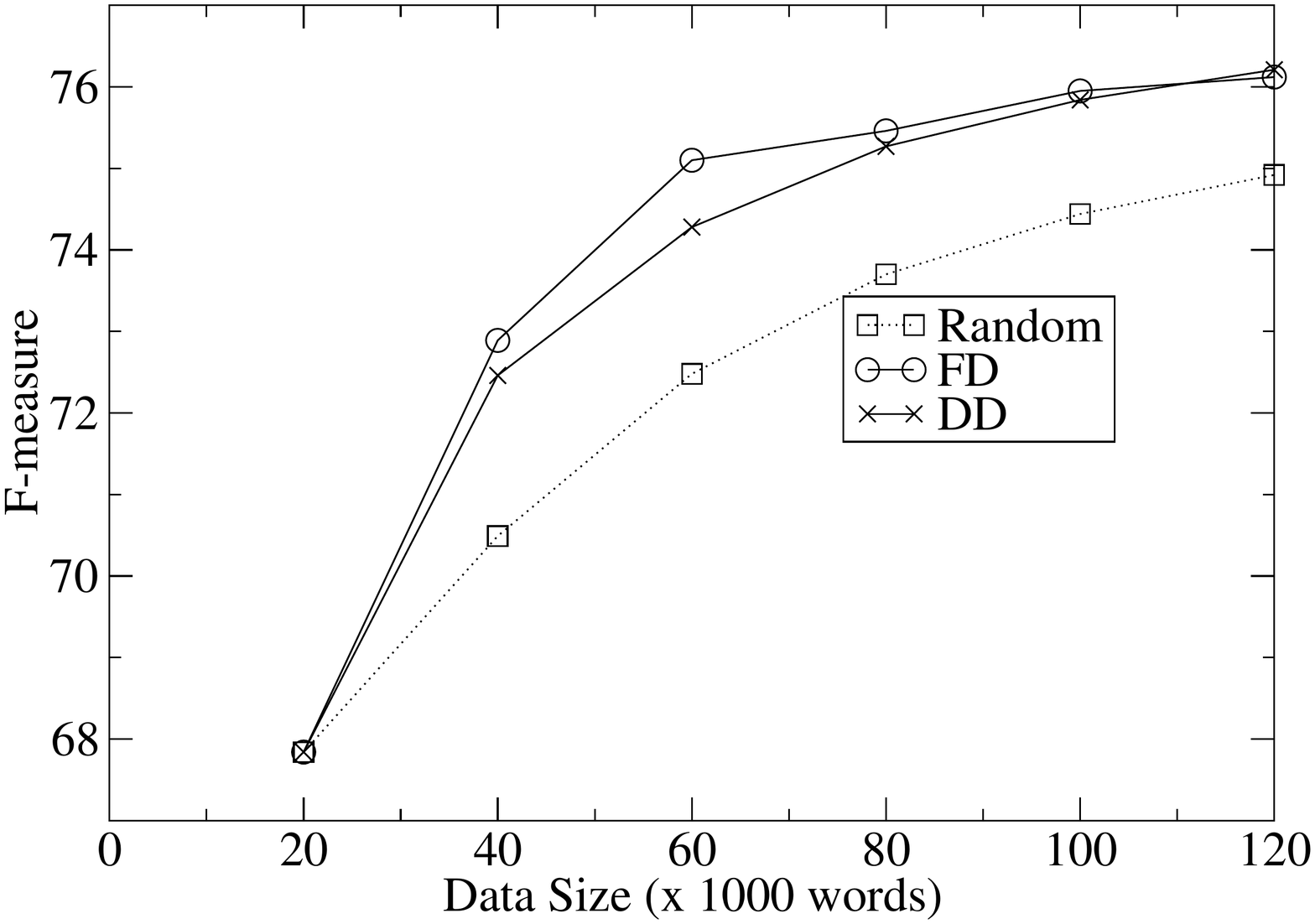}
\caption{Performance for the Confidence Sum metric, computed for both classifier training settings.}
\label{confsum}
\end{minipage}
\hspace{0.1cm} 
\begin{minipage}[b]{0.6\linewidth}
\mbox{}\hspace{-0.2in}
\includegraphics[scale=0.3]{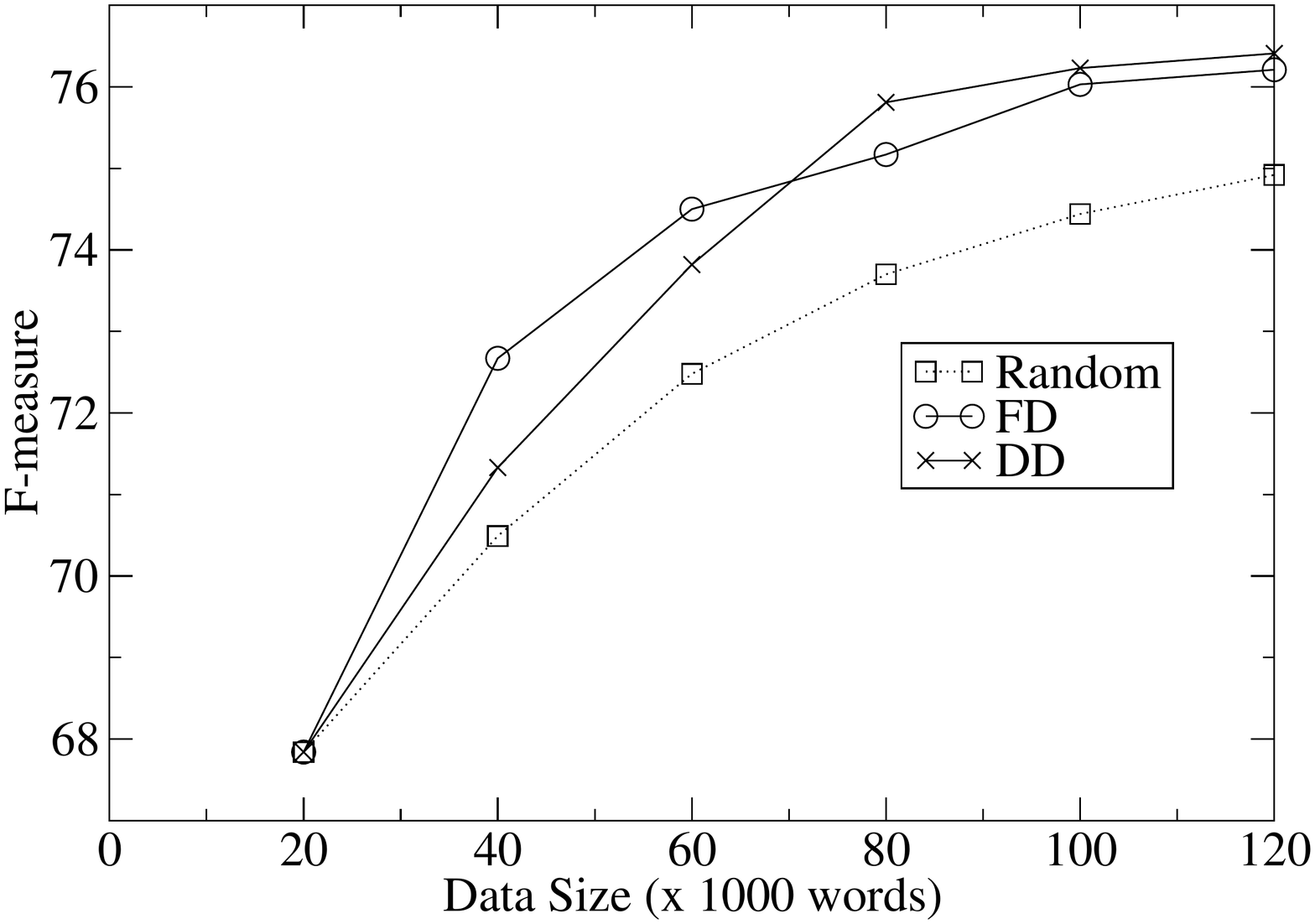}
\caption{Performance for the Confidence Difference metric.}
\label{confdiff}
\end{minipage}
\end{figure}

%

%

\begin{figure}
\mbox{}\hspace{-0.65in}
\begin{minipage}[b]{0.6\linewidth} 
\mbox{}\hspace{-0.2in}
\includegraphics[scale=0.3]{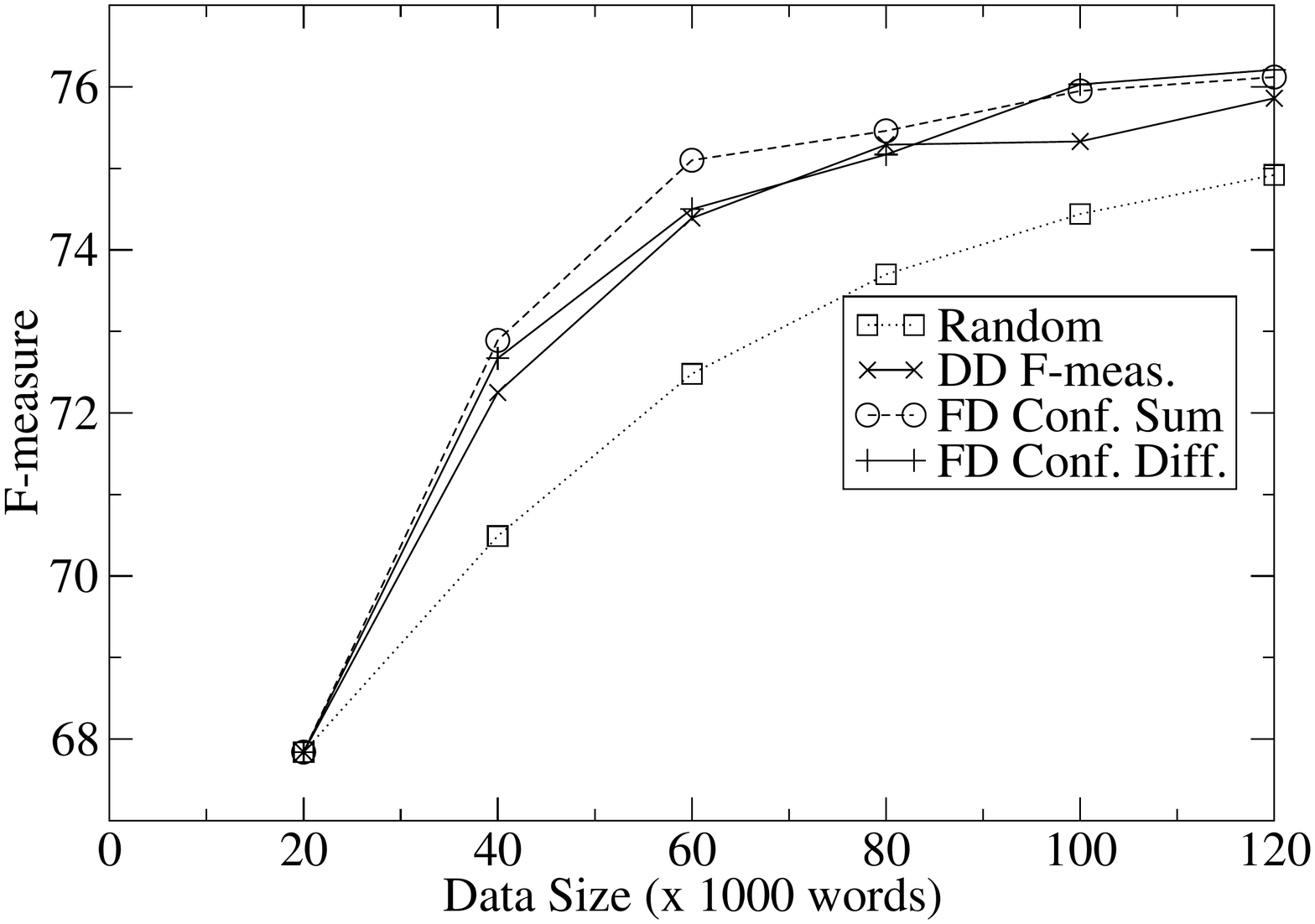}
\caption{Comparing the $3$ best selection strategies.}
\label{best3}
\end{minipage}
\hspace{0.1cm} 
\begin{minipage}[b]{0.6\linewidth}
\mbox{}\hspace{-0.2in}
\includegraphics[scale=0.3]{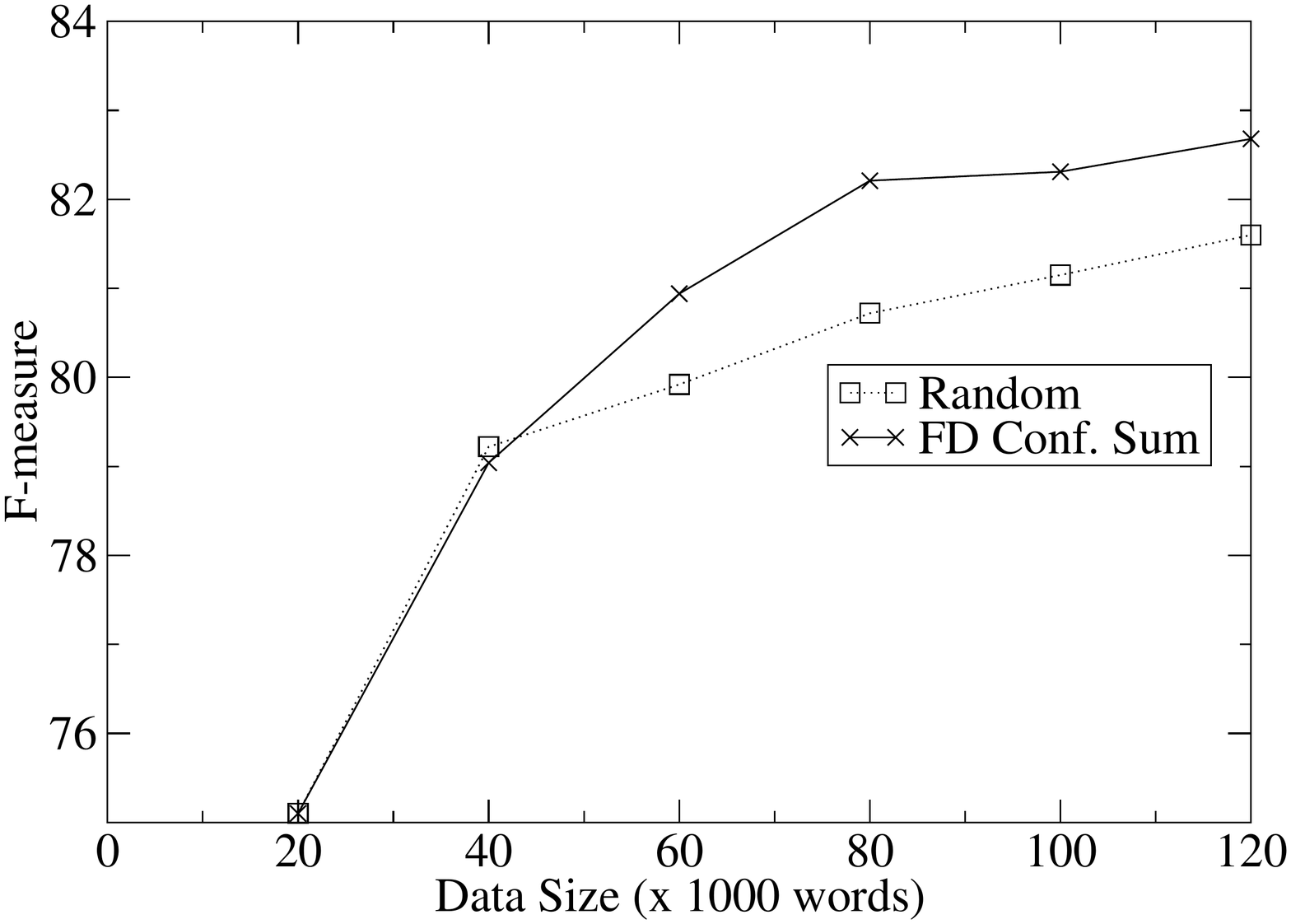}
\caption{Performance of the \emph{PERSON} category.}
\label{person}
\end{minipage}
\end{figure}

%
%


\textbf{Random Selection}. We established a baseline performance by
selecting sentences at random from the unlabeled data pool. We
performed $5$ runs and averaged the numbers from each run. The
performance for the random selection strategy is shown in
Figure~\ref{baseline}, including each run and their average. The
average performance is used as the baseline and shown in later
graphs. The X-axis represents the data size in number of words, and
the Y-axis presents the F-measure on the development set.


\textbf{F-measure and Macro-Averaged F-measure}. Figure~\ref{fmeas}
shows the curves for the F-measure and Macro-Averaged F-measure
metrics compared to the baseline. As we can see, the F-measure metric
works very well. Even at 40K words (after only one step of active
learning), we observe an increase of about 2 absolute F-measure points
with the same amount of labeled data, compared to random sentence
selection. To attain the same performance, the random strategy needs
$1.5$ times as much labeled data. The selection strategy based on
macro-averaged F-measure also performs better than random selection,
but the improvement is smaller than that for the F-measure metric. As indicated, these metrics were computed only for the data-different classifier setting.


\textbf{Confidence Sum}. Figure~\ref{confsum} shows the curves for the
confidence sum metric. This metric is computed for both the data-different and the feature-different setting. As shown in the figure, the
classifiers trained on different features (FD) lead to slightly higher improvement in the
first two steps than the classifiers with the same feature set (DD). Overall, the confidence sum performs much better than random
selection: at 60K words (after only two steps of active learning), the confidence sum strategy attains better performance than the baseline does with twice this amount of labeled data.


\textbf{Confidence Difference}. Figure~\ref{confdiff} shows the results
for the confidence difference metric. Similar to the confidence sum
metric, using the classifiers trained on different features (FD) leads to better improvements in
the first two steps. At 60K words, the system achieves the same performance as the random strategy at 100K words.


\textbf{The Best 3 Strategies}. Figure~\ref{best3} compares the curves
for the $3$ best strategies - FD confidence sum ,
FD confidence difference, and DD F-measure. The
FD confidence sum strategy outperforms the other two strategies in the first three active
learning steps. At the fourth step, the FD confidence difference strategy catches up. At 60K words, the FD confidence sum strategy achieves the performance of the random strategy at about 130K words. This indicates a reduction of over 50\% in the amount of annotated data and demonstrates the effectiveness of active learning for making annotation more efficient.

\begin{figure}
\mbox{}\hspace{-0.65in}
\begin{minipage}[b]{0.6\linewidth} 
\mbox{}\hspace{-0.2in}
\includegraphics[scale=0.3]{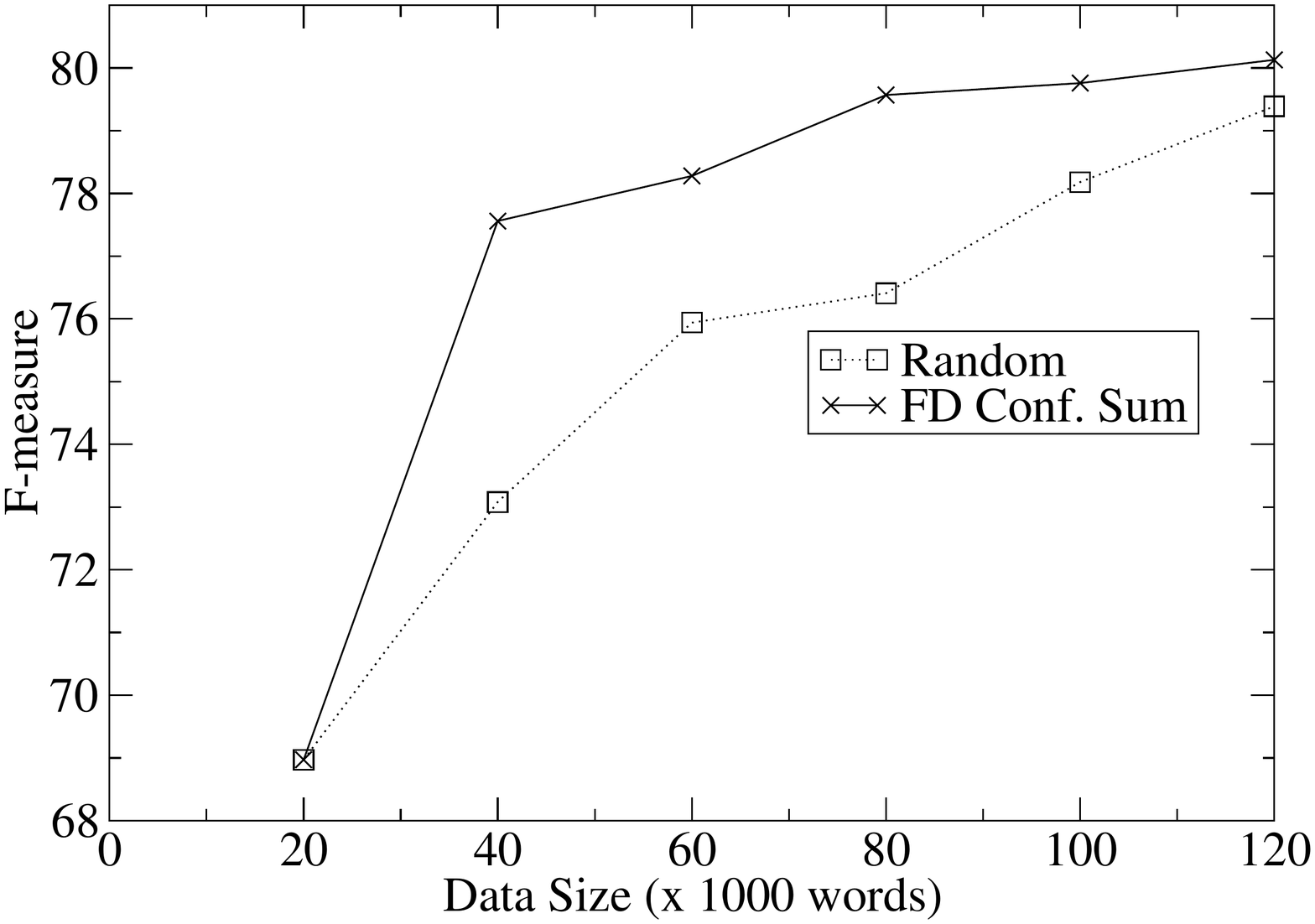}
\caption{Targeting named mentions.}
\label{named}
\end{minipage}
\hspace{0.1cm} 
\begin{minipage}[b]{0.6\linewidth}
\mbox{}\hspace{-0.2in}
\includegraphics[scale=0.3]{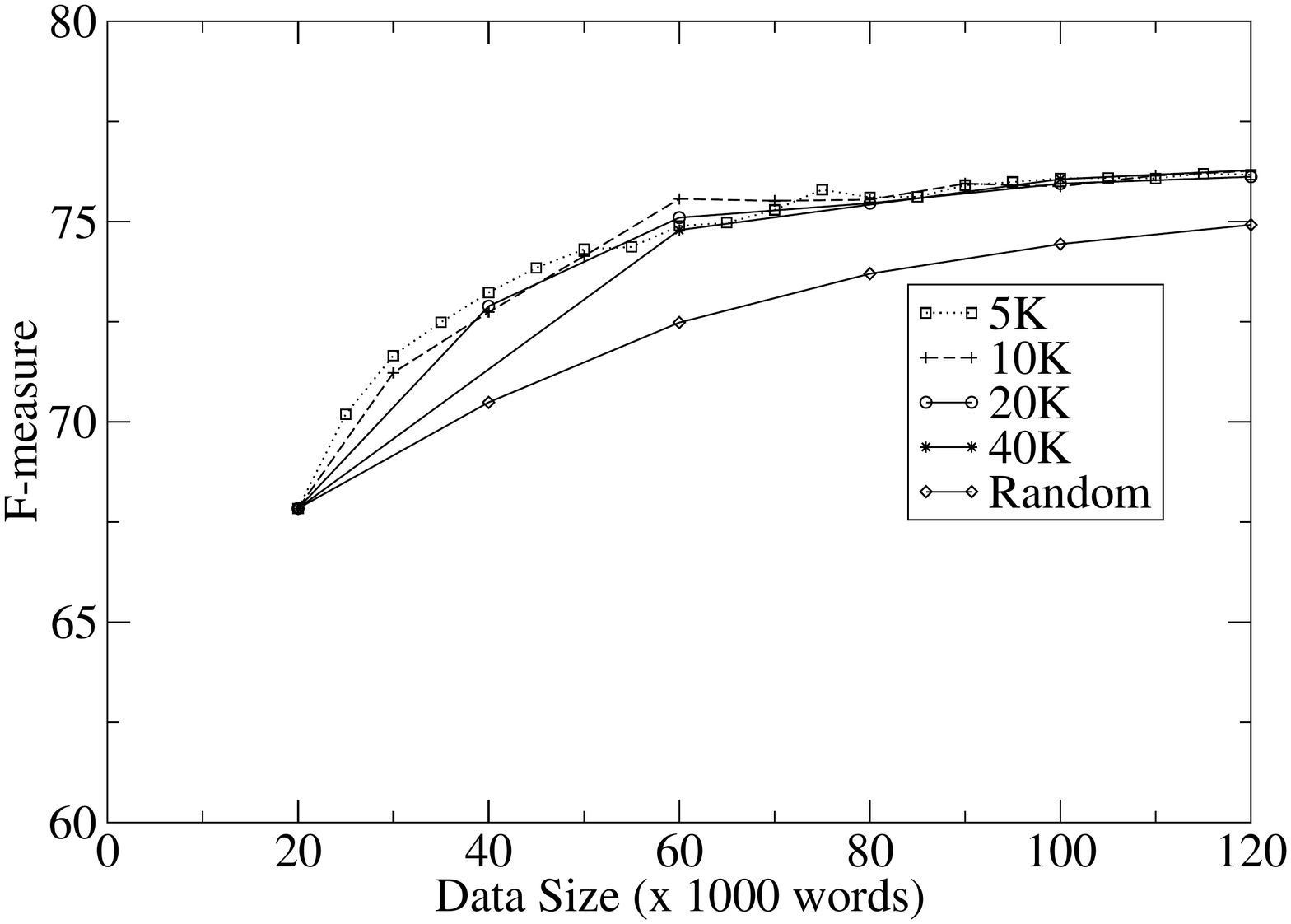}
\caption{Effect of step size on confidence sum metric.}
\label{stepsize}
\end{minipage}
\end{figure}

%

\subsection{Focusing on specific types of mentions}

As we have described in the previous section, our system allows us to weight the different types and levels of mentions differently in
order to focus our learning on specific mention types and levels.  We provide the results of two such experiments: one focusing on the PERSON category and one focusing on named mentions.

First, we target the learning to mentions of type PERSON. To achieve this, we weight the PERSON category twice as much as any of
the other mention categories. The results are shown in Figure~\ref{person}. Note that the F-measure along the y-axis in the graph is the performance for the PERSON category.


The second experiment was for named mentions. For this experiment, we set the weights to all other levels to $0$. The result is an active learner specifically tuned to maximizing the performance of the system on named mentions. Figure~\ref{named} shows the results for this experiment. This plot clearly illustrates the advantages of active learning. We see a $5$-point F-measure increase with only one step of learning, which could otherwise have only been achieved by annotating almost $2.5$ times as much data.

\subsection{Effect of step size}

The size of the step used in the learning needs to be optimized. On the one hand, getting higher performance gains from smaller amounts of labeled data to be added at each step represents a huge savings in the annotation task; on the other hand, retraining active learners at each step takes time and resources. We need to balance the cost of retraining active learners and the gains from smaller batches.

We experimented with different step sizes for the feature-different (FD) confidence-sum metric. The results are shown in Figure~\ref{stepsize}. From that graph, it seems that a step size of $10000$ words might have provided the best balance between annotation and performance. Step sizes higher than $20000$ do not seem to have the right granularity for effective learning.

\section{Conclusions}

We conducted several active learning experiments for the task of detecting mentions of entities in human transcripts of spontaneous conversational speech.  Specifically, we proposed and compared a variety of sentence selection strategies for active learning. The best strategy uses the sum of the confidence values of a pair of classifiers trained with different feature sets. Compared to random sentence selection, this strategy required 50\% of the data to achieve the same performance. For named mentions, the strategy required only 42\% of the data needed by random selection to achieve the same performance.

In the future, we would like to test our active learning strategies on
data from a different domain and data in languages other than English,
such as the ACE mention annotation corpus. We would also like to explore active
learning for co-reference and relation annotation.

\section*{Acknowledgment}
This project was part of a research effort funded by NSF. 
Any opinions, findings and conclusions or recommendations 
expressed in this material are those of the authors and do not 
necessarily reflect the views of the NSF.

\end{document}